\documentclass[conference]{IEEEtran}
\usepackage{cite}
\usepackage[T1]{fontenc}
\usepackage{threeparttable}
\usepackage[table]{xcolor}
\usepackage[latin9]{inputenc}
\PassOptionsToPackage{draft}{hyperref}
\usepackage[bookmarks=false]{hyperref}
\PassOptionsToPackage{bookmarks=false}{hyperref}
\usepackage[output-decimal-marker={.}]{siunitx}
\PassOptionsToPackage{bookmarks=false}{hyperref}
\usepackage{array}
\PassOptionsToPackage{draft}{hyperref}
\usepackage[left=0.64in,right=0.64in,top=0.98in,bottom=0.98in]{geometry}
\usepackage{subfig}
\renewcommand{\arraystretch}{1.2}
\newcolumntype{C}{>{$}c<{$}}
\usepackage{booktabs}
\usepackage{caption}

\usepackage{graphics}
\newsavebox{\foobox}

\usepackage{bm}
\usepackage{makecell}
\usepackage{tikz}
\usetikzlibrary{shapes.arrows}
\usepackage{color}
\usepackage{xcolor}
\usepackage{array}
\usepackage{booktabs}
\usepackage{textcomp}
\usepackage{multirow}
\usepackage{mathtools}

\usepackage{url}
\usepackage{pifont}
\usepackage{amsfonts}
\usepackage{indentfirst}
\usepackage{amsthm}
\usepackage{amssymb}
\usepackage[export]{adjustbox}
\usepackage{verbatim}
\usepackage{graphicx}
\usepackage{mathrsfs} 
\DeclareMathAlphabet{\mathpzc}{OT1}{pzc}{m}{it}
\usepackage{booktabs}

\usepackage{color}

\usepackage[ruled,linesnumbered]{algorithm2e}
\definecolor{seagreen}{rgb}{0.18, 0.55, 0.24}
\SetAlFnt{\small\color{black}\tt}
\LinesNumbered

\usepackage{graphicx}
\usepackage{placeins}

\usepackage{subfig}

\newcounter{defcounter}
\usepackage[shortlabels]{enumitem}
\usepackage{cite}
\usepackage[ruled]{algorithm2e}
\mathchardef\period=\mathcode`.
\DeclareMathSymbol{.}{\mathord}{letters}{"3B}
\usepackage{tikz}
\tikzstyle{io} = [fill=black,inner sep=2pt,circle]

\graphicspath{ {images/} }

\usetikzlibrary{arrows,positioning}
\usetikzlibrary{shapes.geometric,positioning,shapes,arrows,calc, decorations.pathreplacing, arrows.meta}

\everymath{\displaystyle}
\usetikzlibrary{backgrounds}

\usepackage{pifont}

\makeatletter
\def\endthebibliography{%
	\def\@noitemerr{\@latex@warning{Empty `thebibliography' environment}}%
	\endlist
}

\makeatletter
\newcommand*\bigcdot{\mathpalette\bigcdot@{.5}}
\newcommand*\bigcdot@[2]{\mathbin{\vcenter{\hbox{\scalebox{#2}{$\m@th#1\bullet$}}}}}
\makeatother

\makeatletter

\theoremstyle{plain}

\usepackage{cite} 
\usetikzlibrary{shapes,arrows}
\tikzstyle{line}=[draw] 
\usepackage{algorithmic}
\usepackage{graphicx}
\usepackage{enumitem}

\usepackage{url}
\usepackage{soul}
\usepackage{xcolor}

\makeatother
\usepackage[english]{babel}
\providecommand{\theoremname}{Theorem}

\usepackage{sansmath}

\begin{document}

\title{On Sharing Models Instead of Data using Mimic learning for Smart Health Applications}

\author{\IEEEauthorblockN{
 Mohamed Baza\IEEEauthorrefmark{1},
   Andrew Salazar\IEEEauthorrefmark{2},
  Mohamed Mahmoud\IEEEauthorrefmark{1},
  Mohamed Abdallah\IEEEauthorrefmark{3},
  Kemal Akkaya\IEEEauthorrefmark{3}
  }

   \IEEEauthorblockA{%
    \IEEEauthorrefmark{1}Department of Computer Science, Tennessee Tech University, Cookeville, TN, USA
  }
  \IEEEauthorblockA{%
    \IEEEauthorrefmark{3}Department of Information and Decision Sciences, California State San Bernardino, San Bernardino, CA, USA
  }
  
  \IEEEauthorblockA{%
    \IEEEauthorrefmark{3}division of Information and Computing Technology, College of Science and Engineering, HBKU, Doha, Qatar
  }
  
    \IEEEauthorblockA{
    \IEEEauthorrefmark{4}Department of Electrical and Computer Engineering, Florida International University, Miami, FL, USA}
\vspace{-0.25in}
}

\maketitle
\IEEEpeerreviewmaketitle

 \begin{abstract}
 
Electronic health records (EHR) systems contain vast amounts of medical information about patients. These data can be used to train machine learning models that can  predict health status, as well as to help prevent future diseases or disabilities. However, getting patients' medical data to obtain well-trained machine learning models is a challenging task. This is because sharing the patients' medical records is prohibited by law in most countries due to patients privacy concerns.  In this paper, we tackle this problem by sharing the models instead of the original sensitive data by using the mimic learning approach. The idea is first to train a model on the original sensitive data, called the teacher model. Then, using this model, we can transfer its knowledge to another model, called the student model, without the need to learn the original data used in training the teacher model. The student model is then shared to the public and can be used to make accurate predictions. To assess the mimic learning approach, we have evaluated our scheme using different medical datasets. The results indicate that the student model mimics the teacher model performance in terms of prediction accuracy without the need to access to the patients' original data records.

	\end{abstract}

	\vspace{-0.1in}
\section{Introduction}


Over the past few years, the adoption of electronic
health records (EHRs) by health care systems have increased significantly~\cite{birkhead2015uses}. According to~\cite{shickel2017deep}, nearly 84\% of hospitals have adopted at least a basic EHR system~\cite{shickel2017deep}. The main goal of EHRs systems is to store detailed patients data such as demographic information, diagnoses,
laboratory tests and results, prescriptions, radiological images,
clinical notes, and more~\cite{arndt2017tethered}.

Recently, the research community increasingly incorporates machine learning algorithms in the EHRs domain~\cite{rajkomar2018scalable}. The primary goal of these algorithms is to develop models that can be used by physicians to predict health status and help prevent future diseases or disabilities. As an example, Google has developed a machine learning algorithm to help identify cancerous tumours on mammograms~\cite{ref:google}. Also, Google's DeepMind Health project~\cite{ref:google2} aims to create the best treatment plans for cancer patients by rapidly analyzing their medical test results and instantly referring them to the right specialist. 

It is well known that data drives machine learning~\cite{obermeyer2016predicting}. As more data is available, as it is more likely for machine learning algorithms to give accurate predictions that doctors can use. However, patient's records maintained by hospitals can not be shared due to patient privacy concerns. In most countries, privacy protection laws have been passed to protect patients data from being shared or leaked. For instance, 
In 1996, the Health Insurance Portability and Accountability Act (HIPAA) title II was enacted in the U.S.A~\cite{2}. One of the primary objectives of this act is to increase the protection of patients' medical records against unauthorized usage and disclosure. These laws prevent hospitals from sharing the medical data records since with detailed person-specific records, sensitive information about patients may be easily revealed by analyzing the shared data. Even if the data is anonymized before sharing it, research has shown that patients could be easily identified by using specific combined information (such as age, address, sex, etc.). For example,~\cite{samarati2001protecting} shows that linking medication records with voter lists can uniquely identify a person's name and his/her medical information. Therefore, \textit{privacy concerns hinder sharing the patients' records, which makes creating well-trained machine learning models a challenging task}.

\begin{figure*}[!t]
		\setlength{\abovecaptionskip}{0.1cm}
		\setlength{\belowcaptionskip}{-0.7cm}
		\centering
		\subfloat[SVM classifier  \label{checkcomptime}]
		{\includegraphics[width=0.27\linewidth]{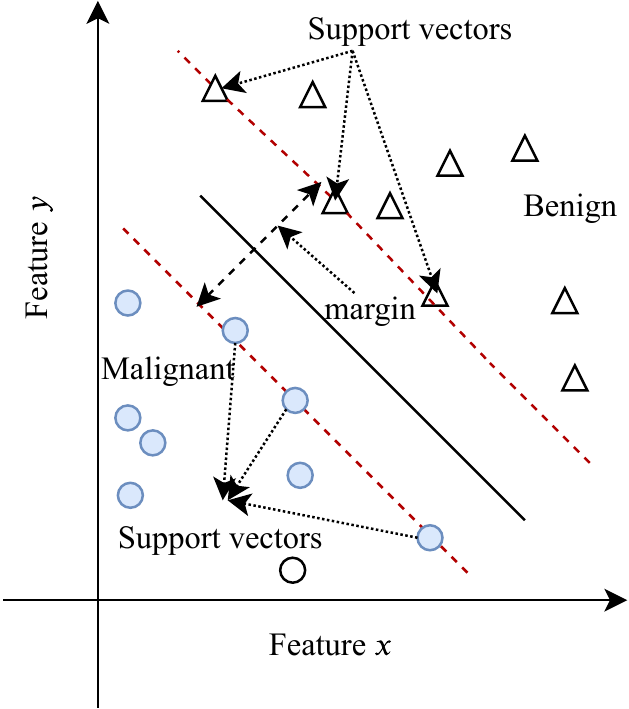}} \hspace{20pt}
		\subfloat[KNN classifier with $k=8$ \label{trajcomptime}]
		{\includegraphics[width=0.27\linewidth]{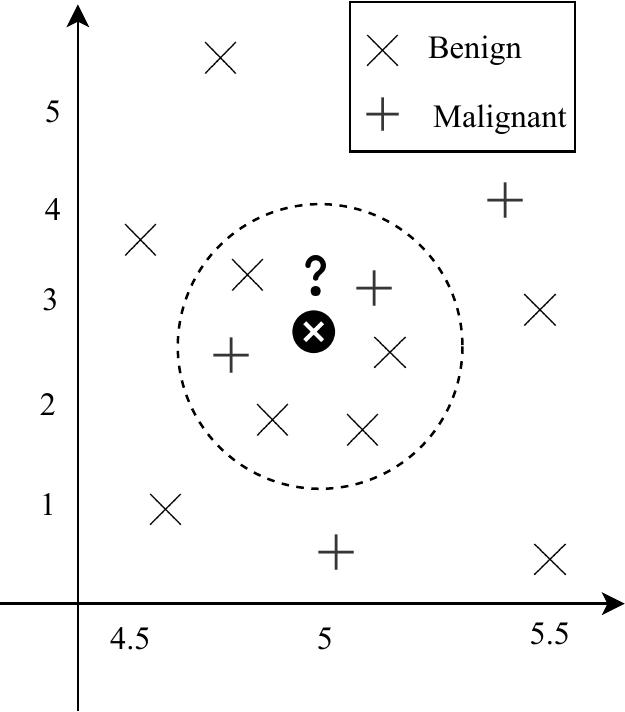}}\hspace{14pt}
		\subfloat[RF classifier \label{forgecomptime}]
		{\includegraphics[width=0.3\linewidth]{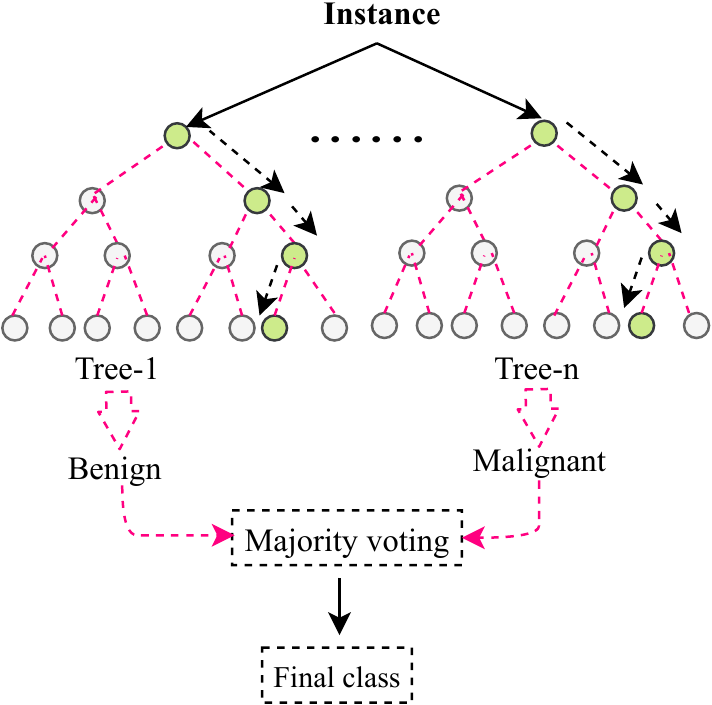}}
	
		\vspace{10pt}
		\caption{Machine learning classifiers used in our scheme.}
		
		\label{fig:computation_new}
	\vspace{10pt}
	\end{figure*}

In this paper, we tackle the aforementioned problem by sharing the machine learning models instead of the data using mimic learning approach. The main idea of mimic learning is to enable the transfer of knowledge from sensitive private EHR records to a shareable model while protecting the original patients data from being shared. In a nutshell, a \textit{teacher model} trained on the original sensitive patients data, is then used to annotate a large set of unlabeled public data. Then, these labeled data are used to train a new model called the \textit{student model}. The student model can be shared to make accurate predictions without the need to share original data or even the teacher model. To give empirical evaluations of the mimic learning approach in EHR domain, we have used three different datasets of patients data. The results indicate that the student model follows the teacher model in making accurate predictions. Moreover, as the teacher accuracy increases, the student model performance follows the teacher on the unseen data.

The remainder of this paper is organized as follows. The background is presented in \ref{Background}. Our proposed scheme is discussed in details in Section~\ref{proposed}. The experimental results are discussed in Section~\ref{exp}. Section~\ref{Related} discusses the previous related work. Our conclusions are presented in Section~\ref{Conclusion}.

\section{Background}
\label{Background}

Our scheme includes the following four machine learning algorithms: Support Vector Machine (SVM), k-Nearest Neighbor (KNN), Random Forests (RF), and Na\"{i}ve Bayes (NB). In this section, we provide an overview for these algorithms.

\begin{itemize}

\item \textit{Support Vector Machine (SVM).} An SVM model is a representation of the examples as points in space, mapped so that the examples of the separate categories are divided by a clear gap that is as wide as possible\cite{Cortes:1995:SN:218919.218929}. In the SVM model, decision hyperplanes are formed based on identified support vectors to create a separation gap to divide two class instances with the maximal margin, as shown in Fig.~\ref{checkcomptime}. This is done by finding the hyperplane that has the most significant fraction of points of the same class on the same plane.


\item \textit{K-Nearest Neighbor (KNN).} The KNN classifier is a semi-supervised learning algorithm that identifies the classes of {\sansmath$K$} nearest neighbors of a given instance based on the majority class obtained~\cite{shruthi2019review}. The algorithm takes as an input the training examples {\sansmath $\{x_i,y_i\}$}, where {\sansmath$x_i$} is the attribute-value representation of the examples, and {\sansmath$y_i$} is the class label (e.g., benign or malignant), and a testing point {\sansmath$x$} that we want to classify. The KNN first computes the distance {\sansmath$D(x,x_i)$} to every training example {\sansmath$x_i$}. Then, it selects the {\sansmath$k$} closest instances \{\sansmath$x_{i1},\ldots,x_{ik}$\} and their labels \{\sansmath$y_{i1},\ldots,y_{ik}$\}. Finally, it output the class {\sansmath$y$} which is the most frequent of \{\sansmath$y_{i1},\ldots,y_{ik}$\}. 
As illustrated in Fig.~\ref{trajcomptime}, the point to
be classified is (5, 2.45), which is shown with $\otimes$. When
applying KNN algorithm with {\sansmath$k = 6$} using
Euclidean distance computation, the result is shown with a dotted circle. There are two possible cases: \textit{Malignant} class 
with two instances and \textit{Benign} class with five instances.
The algorithm classifies the mark $'X'$ to the malignant class
 since it represents the majority of data within the radius.

\item \textit{Random Forests (RF)}:  A random forest is a classifier that consists of multiple decision trees, each of which provides a \textit{vote} for a specific class~\cite{Breiman2001}. Combining a large number of trees in a random forest leads to more reliable predictions, while a single decision tree may overfit the data. As illustrated in Fig.~\ref{forgecomptime}, an instance is labeled to the class that is selected by the majority of the trees votes in the forest. We adopted RF for its high capability of avoiding overfitting problem.

\begin{figure*}[ht]
\begin{minipage}[b]{0.45\linewidth}
\centering
\includegraphics[width=0.93\textwidth]{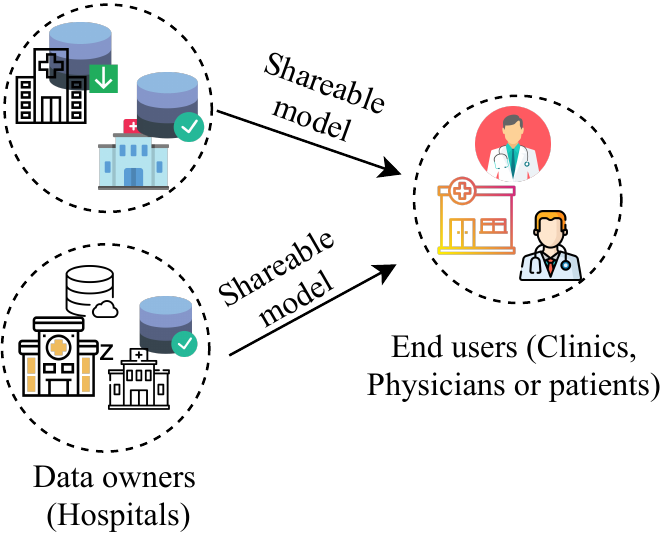}
\vspace{0.4cm}
 \caption{Illustration of the system model under consideration} 
\label{fig1: System} 
\end{minipage}
\hspace{1cm}
\begin{minipage}[b]{0.45\linewidth}
\centering
\includegraphics[width=0.9\textwidth]{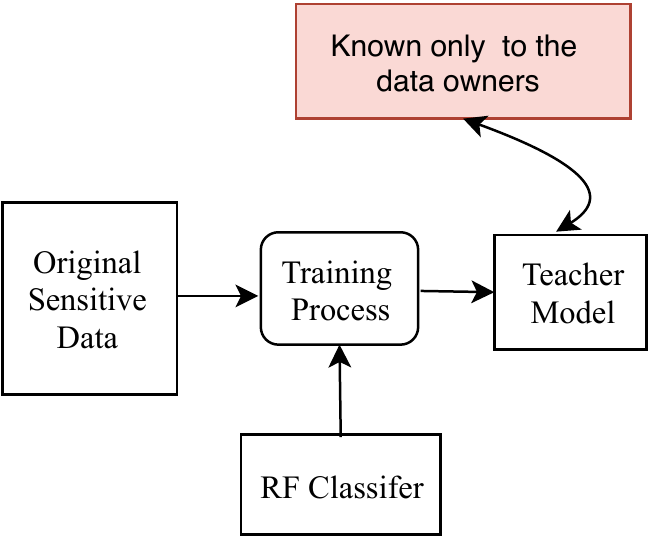}
\vspace{0.3cm}
\caption{Teacher model generation process} 
    \label{fig: mimic}
\end{minipage}
\begin{minipage}[b]{0.45\linewidth}
\centering
\includegraphics[width=0.9\textwidth]{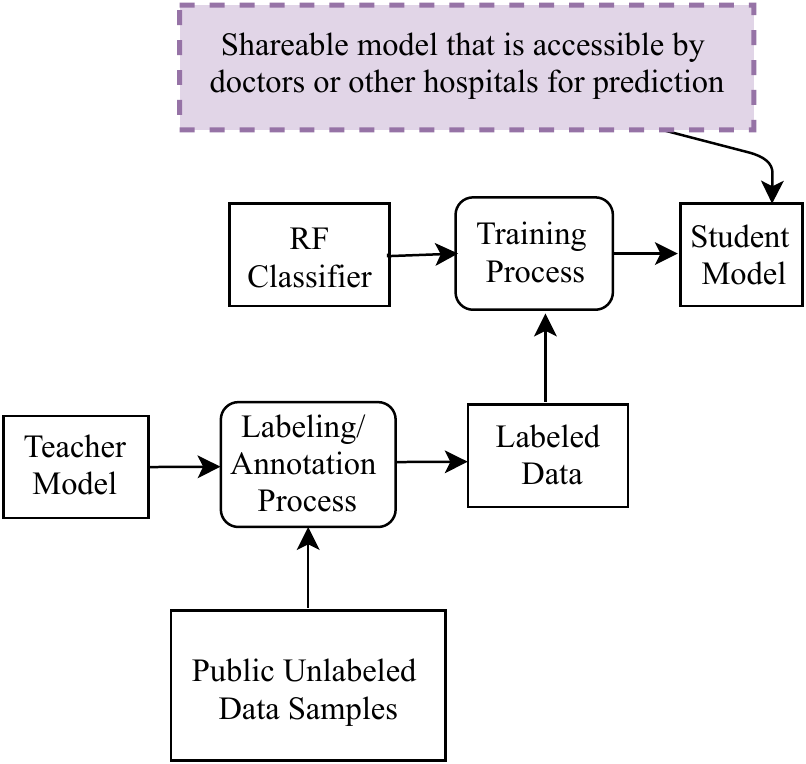}
\vspace{0.3cm}
\caption{Student model generation process} 
    \label{fig: student}
\end{minipage}
\hspace{1cm}
\begin{minipage}[b]{0.45\linewidth}
\centering
\includegraphics[width=0.8\textwidth]{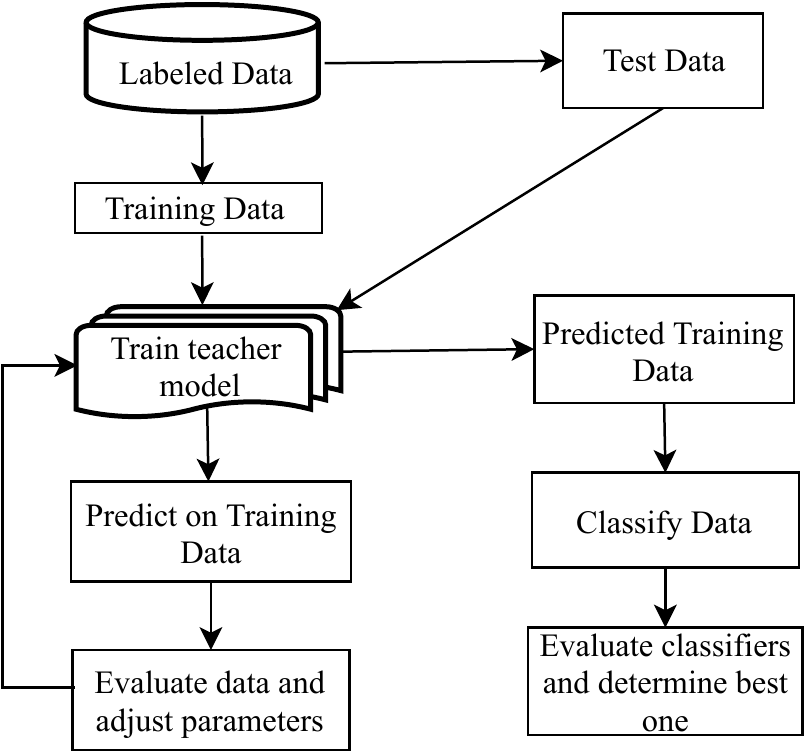}
\vspace{0.3cm}
  \caption{Illustration of training process in the teacher/student
models generation.} 

\label{fig: training}
\end{minipage}
\end{figure*}

\item \textit{Naive Bayes (NB).} Is a statistical classifier which
assumes no dependency between attributes. It attempts to
maximize the posterior probability in determining the class by assuming that the presence (or
absence) of a particular feature of a class is
unrelated to the presence (or absence) of any other
feature. For example, a fruit may be considered to
be an apple if it is red, round, and about 4" in
diameter. Even though these features depend on the
existence of the other features, a naive Bayes
classifier considers all of these properties to
independently contribute to the probability that this
fruit is an apple.

\end{itemize}

\section{Proposed Scheme}
\label{proposed}
In this section, we describe our proposed scheme. We first describe the system architecture, followed by the generation processes of teacher and student models.

\subsection{System Architecture}
As illustrated in Fig.~\ref{fig1: System}, the system architecture includes two main entities namely, \textit{data owners} and \textit{end users}. Data owners can be large hospitals that own the patients medical data records. They also do not want to share these data with others due to patients privacy concerns. They also responsible for generating the teacher and student models. The end users can be the physicians or even other hospitals that want to make use of machine learning models early prediction of diseases and creating better treatment plans.


\subsection{Teacher Model Generation}

The first step is to generate the teacher model. As illustrated in Fig. \ref{fig: mimic}, data owners use their original sensitive data to train a set of machine learning classifiers to obtain the corresponding models. Then, theses models are then compared according to their performance, and the most accurate model is selected to be used as the teacher model.

The training process is illustrated in Fig.~\ref{fig: training}. We first begin with splitting the labeled data into training and testing data. The training data is then used for training the teacher models, evaluating the results, and changing the weights and biases to more accurately predict the data. Once the training process is done, the test data are classified, and the best classifier is selected to be the teacher model.
 
Notice that the teacher model is kept private and can not be shared to the public. This is because if any adversary has access to the teacher model, this can lead to learning sensitive genomic information about individuals using some kind of model inversion attack~\cite{fredrikson2014privacy}.

\subsection{Student Model Generation}

The generation of the student model starts after generating the teacher model. The process of computing the student model is illustrated in Fig.~\ref{fig: student}. The first step is the \textit{labelling/annotating} process. In this step,  the teacher model is used to label (or annotate) an unlabeled public dataset to generate a new labeled training data.

After the labelling/annotating process is performed, the generated labeled data is used in training and selection of the student model as illustrated in Fig.~\ref{fig: mimic}. The training process is similar to Fig.~\ref{fig: training}. The major difference between training the student and teacher models is that the training process of the student model is basically a knowledge transfer process. In other words, the knowledge that the teacher model has gained from the sensitive patients data is transferred to the student model by using publicly available data produced in the annotation process.

Finally, once the student model is obtained, its performance needs to be evaluated with the teacher model to ensure its accuracy before sharing it.



\section{Experiments}
\label{exp}
In this section, we first explain the data and the process used
to evaluate our scheme. We also present and discuss the results of this
evaluation.

\subsection{Datasets Description}
 We used three different datasets in our experiments. Table~\ref{table:compare}, gives description overview for datasets. 
 
 \begin{itemize}

     \item \textbf{Breast-cancer dataset} is a property of UCI's machine learning repository~\cite{UCI} which contains breast cancer diagnostic of more than 699 patients. In this dataset, features are computed from digitized images of a fine needle aspirate of a breast mass. The images describe the characteristics of the cell nuclei present in the image
    
     \item \textbf{Cardiovascular dataset} is obtained from kaggle data science~\cite{Kaggle}. The dataset contains detailed medical examination of about 70000 patients having problems with cardiovascular diseases. The data includes factual information (age, gender, height, weight), examination results (systolic blood pressure, diastolic blood pressure, cholesterol, glucose levels) and subjective information that the patient provides (smoker, alcohol intake, activity level). 
    
     \item \textbf{Heart disease dataset} is a property of UCI's machine learning repository~\cite{UCI2}. The data includes information such as age and sex, test results such as resting blood pressure, serum cholesterol, fasting blood sugar, and resting electrocardiographic results, and subjective information such as chest pain type. 
    
     \end{itemize}

\begin{table}[!t]
 \centering
\renewcommand{\arraystretch}{1.35}
	\caption{Datasets used in the evaluations.}
	\label{table:compare}
	\begin{tabular}{l|c|c}
		\hline 	\Xhline{3\arrayrulewidth}
	Disease	& No. of samples & No. of features \\ 
		\hline	
		Breast cancer	& 699 & 18 \\\hline
		Cardiovascular & 70000 & 11 \\\hline
		Heart disease & 303 &14 \\\hline
	\Xhline{3\arrayrulewidth}
	\end{tabular}
\end{table}

\subsection{Performance Metrics}

\begin{table*}[!t]

        \centering
        \caption{The experiment results of the teacher and student models for the \textit{breast-cancer} data.}
        \label{table1}
        \begin{tabular}{|c|c|c|c|c|c|c|c|c|c|}
        
            \cline{1-9}
             & \multicolumn{4}{|c|}{\cellcolor{blue!20}Teacher model}& \multicolumn{4}{|c|}{\cellcolor{red!20}Student model}\\
            \cline{2-9}
        & Precision & Recall & F1-Score& Accuracy & Precision & Recall & F1-Score& Accuracy\\
            \hline
    
    SVM   & 0.849 & 0.862 & 0.822 & 0.8546 & 0.945 & 0.941 & 0.935 & 0.9382\\ \hline

    KNN  & 0.969 & 0.967 & 0.964 & 0.9673 & 0.962 & 0.958 & 0.956 & 0.9636\\ \hline
    
    \textbf{\textcolor{red}{RF}}   & \textbf{\textcolor{blue}{0.966}} & \textbf{\textcolor{blue}{0.967}} & \textbf{\textcolor{blue}{0.965}} & \textbf{\textcolor{blue}{0.9697}} & \textbf{\textcolor{red}{0.969}} & \textbf{\textcolor{red}{0.966}} & \textbf{\textcolor{red}{0.964}} & \textbf{\textcolor{red}{0.9673}}\\ \hline

    NB  & 0.963 & 0.961 & 0.961 & 0.9618 & 0.945 & 0.94 & 0.942 & 0.94\\ \hline
        \end{tabular}
         
    \end{table*}

\begin{table*}[!t]
        \centering
\label{table2}
        \caption{The experiment results of the teacher and student models for the \textit{heart-disease} data.}
    
       \label{table2}
        \begin{tabular}{|c|c|c|c|c|c|c|c|c|c|}
         
            \cline{1-9}
             & \multicolumn{4}{|c|}{\cellcolor{blue!20}Teacher model}& \multicolumn{4}{|c|}{\cellcolor{red!20}Student model}\\
            \cline{2-9}
        & Precision & Recall & F1-Score& Accuracy & Precision & Recall & F1-Score& Accuracy\\
            \hline

    SVM   & 0.8633 & 0.84 & 0.836667 & 0.8387 & 0.826 & 0.813 & 0.816 & 0.8172\\ \hline

    KNN  & 0.86 & 0.85 & 0.85 & 0.8495 & 0.807 & 0.783 & 0.78 & 0.7850\\ \hline
    
    \textbf{\textcolor{red}{RF}}   & \textbf{\textcolor{blue}{0.903}} & \textbf{\textcolor{blue}{0.893}} & \textbf{\textcolor{blue}{0.893}} & \textbf{\textcolor{blue}{0.8925}} & \textbf{\textcolor{red}{0.843}} & \textbf{\textcolor{red}{0.84}} & \textbf{\textcolor{red}{0.84}} & \textbf{\textcolor{red}{0.8387}}\\ \hline

    NB  & 0.8 & 0.793 & 0.793 & 0.8172 & 0.8433 & 0.84 & 0.8433 & 0.8387\\ \hline
        \end{tabular}
    \end{table*}

\begin{table*}[!t]
\label{table3}
        \centering
            \caption{The experiment results of the teacher and student models for the \textit{cardiovascular-disease} data.}
    
         \label{table3}
       
        \begin{tabular}{|c|c|c|c|c|c|c|c|c|c|}
         
            \cline{1-9}
             & \multicolumn{4}{|c|}{\cellcolor{blue!20}Teacher model}& \multicolumn{4}{|c|}{\cellcolor{red!20}Student model}\\
            \cline{2-9}
        & Precision & Recall & F1-Score& Accuracy & Precision & Recall & F1-Score& Accuracy\\
            \hline
   
    SVM   & 0.6333 & 0.6833 & 0.66 & 0.6420 & 0.63 & 0.61 & 0.59667 & 0.6107\\ \hline

    KNN  & 0.71667 & 0.70667 & 0.70667 & 0.7083 & 0.61 & 0.6 & 0.5933 & 0.6004\\ \hline
   
   \textbf{\textcolor{red}{RF}}   & \textbf{\textcolor{blue}{0.73}} & \textbf{\textcolor{blue}{0.7267}} & \textbf{\textcolor{blue}{0.7267}} & \textbf{\textcolor{blue}{0.7264}} & \textbf{\textcolor{red}{0.6867}} & \textbf{\textcolor{red}{0.68}} & \textbf{\textcolor{red}{0.6767}} & \textbf{\textcolor{red}{0.6798}}\\ \hline

    NB  & 0.6633 & 0.59 & 0.5367 & 0.5883 & 0.6867 & 0.6833 & 0.6833 & 0.6842\\ \hline
        \end{tabular}
    \end{table*}

For the performance evaluation in the experiment. First, we
denote \textsl{TP}, \textsl{FP}, \textsl{TN}, and \textsl{FN} as true positive (the number of instances correctly predicted true on individuals who are diagnosed with the disease), false positive (the number of instances incorrectly predicting true individuals who are not diagnosed with a disease), true negative (the number of instances correctly predicted false on individuals who are diagnosed with the disease), and false negative (the number of instances incorrectly predicted as false on individuals diagnosed with the disease), respectively. Then, we define the following key performance metrics used in
our evaluation process:

\begin{itemize}
  
    \item \textsl{Accuracy} is the ratio of correctly predicted instances (true negatives and true positives) to the overall number of patients.
    
 \begin{equation}
\textsl{Accuracy}=\frac{\textsl{TP}+\textsl{TN}}{\textsl{TP}+\textsl{TN}+\textsl{FP}+\textsl{FN}}\label{eq:1-1}
\end{equation}

    \item \textsl{Precision} is the proportion of correct positive classifications (true positives) from the cases that are predicted as positive.

        \begin{equation}
\textsl{Precision}=\frac{\textsl{TP}}{\textsl{TP}+\textsl{FP}}\label{eq:1-2}
\end{equation}

    \item \textsl{Recall} is the proportion of correct positive classifications (true positives) from the cases that are \textit{actually} positive for a specific disease.
        
        \begin{equation}
        \textsl{Recall}=\frac{\textsl{TP}}{\textsl{TP}+\textsl{FN}}\label{eq:1-3}
        \end{equation}

\begin{table*}[!ht]
        \centering
      
        \caption{Comparison of the results of the the teacher and student models using the RF classifier for the \textit{breast cancer, heart, and cardiovascular diseases data.}}

        \label{multiprogram}
        \begin{tabular}{|c|c|c|c|c|c|c|c|c|c|c|c|c|c}
         
            \cline{1-13}
             & \multicolumn{4}{|c|}{\cellcolor{green!20} Breast cancer}& \multicolumn{4}{|c|}{\cellcolor{yellow!70}Heart disease}& \multicolumn{4}{|c|}{\cellcolor{magenta!20} Cardiovascular disease}\\
            \cline{2-13}
        & Precision & Recall & F1-Score& Accuracy & Precision & Recall & F1-Score& Accuracy& Precision & Recall & F1-Score& Accuracy\\
            \hline
        Teacher    & 0.97 & 0.97 & 0.97 & 0.96 & 0.90 & 0.89 & 0.89& 0.89&0.73&0.72&0.72&0.72\\
            \hline
      Student   & 0.97 & 0.97 & 0.964 & 0.96 & 0.843 & 0.84 & 0.84 & 0.83&0.68&0.68&0.67&0.67\\
            \hline
     
        \end{tabular}
    \end{table*}

\begin{figure*}[!ht]
		\setlength{\abovecaptionskip}{0.1cm}
		\setlength{\belowcaptionskip}{-0.7cm}
		\centering
		\subfloat[Breast cancer. \label{roc1}]
		{\includegraphics[width=0.3\linewidth]{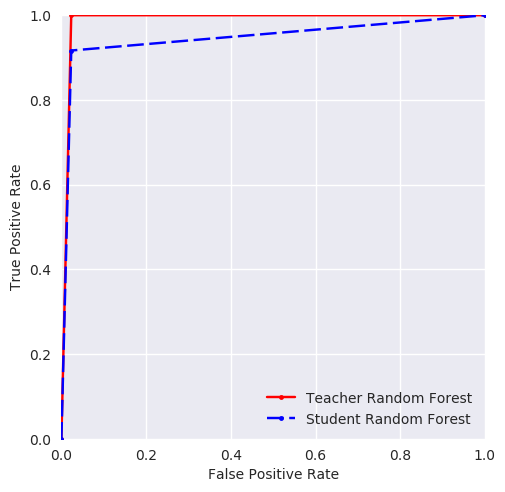}}
		\subfloat[Heart disease. \label{roc2}]
		{\includegraphics[width=0.3\linewidth]{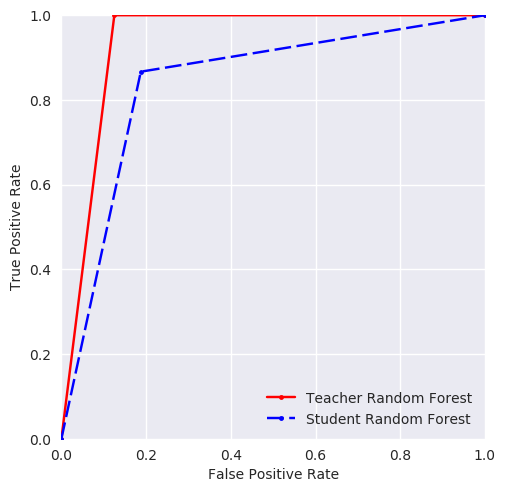}}
		\subfloat[Cardiovascular disease.  \label{roc3}]
		{\includegraphics[width=0.3\linewidth]{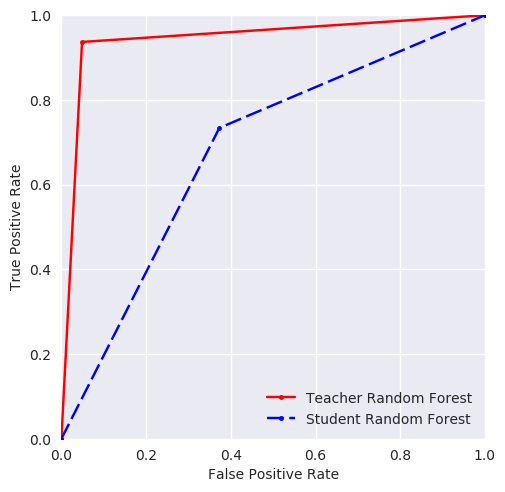}}
		\caption{RoC curves comparison using the RF classifier for both teacher and the student models for all diseases used in our experiment.}
		
		\label{roc}
		\vspace{5pt}
	\end{figure*}

    \item \textsl{F1-Score} is the weighted harmonic mean of the precision and recall. We used F1-score to represent the overall performance of the classifier.

        \begin{equation}
\textsl{F1-Score}=\frac{2 \times ( \textsl{Precision} \times \textsl{Recall })}{\textsl{Precision}+\textsl{Recall}}
\end{equation}

   \end{itemize}


Beside the aforementioned evaluation metrics, we
use receiver operating characteristic (ROC) curve to evaluate the pros and cons of both the teacher and student models for different diseases. The ROC curve shows the trade-off between the true positive rate (\textsl{TPR}) and the false positive rate (\textsl{FPR}),
where the TPR and FPR are defined as follows:

$$
\textsl{TPR}=\frac{\textsl{TP}}{\textsl{TP}+ \textsl{FN}} \quad,\quad \textsl{TFR}=\frac{\textsl{FP}}{\textsl{FP}+\textsl{TN}}
$$

As the ROC curve is closer to the upper left corner of the graph, as
the model performs better. The AUC is the area under the curve.
When the area is closer to one, the model is better. 


\subsection{Results and Discussion}

To evaluate the effectiveness of the mimic learning approach, we used different classifiers mentioned in Sec.~\ref{Background} namely, SVM, KNN, RF and NB. We also used k-fold cross-validation (cv) to reduce the chances of getting biased testing datasets. The idea of the $k$-fold CV is to divide the training data into $k$ equal portions. Then, the model is trained on the remaining $k-1$ folds and the remaining folds are used for evaluation. In all classifiers, we selected the $k$ parameter to be ten \cite{Han:2011:DMC:1972541}.

Each of the four classifiers is used to generate the teacher and student models in each disease. Tables.~\ref{table1},~\ref{table2}, and~\ref{table3} give the results of evaluating the four classifiers for breast cancer, heart, and cardiovascular diseases, respectively. It is clearly seen that the RF classifier outperforms other classifiers when it is evaluated on the original sensitive data in all diseases. We also observe that SVM gives the lowest accuracy in case of the breast cancer dataset, and NB shows the lowest performance in case of heart and Cardiovascular datasets. Therefore, we selected the \textit{RF classifier} as the best teacher model. Similarly, in case of student models, the results clearly indicate that the RF classifier performs the best among this set of classifiers. In addition, the SVM shows the lowest accuracy in the breast cancer dataset, while the KNN gives the lowest accuracy in heart and cardiovascular diseases. Therefore, the RF model is chosen as the best student model.

Then, we compare the performance of the teacher and student models using the RF classifier on test data. The results of the comparison of the different diseases are given in Table~\ref{multiprogram}. The results indicate that the performance of the teacher and student is nearly similar to each other. These results confirm our assertion that unlabeled data trained by a teacher model can be used to transfer knowledge to a student model without revealing data that is considered sensitive.

In another part of our evaluations, we used the RoC curves to visualize the performance of both the teacher and student models, as shown in Fig.~\ref{roc}. The student model has almost identical performance to the teacher model in breast cancer and heart diseases. However, in cardiovascular disease, the gap between student model and teacher model grows more apparently. This is because the teacher model's accuracy is lowest in all three EHRs datasets; thus the student model also follows the teacher model, causing the gap between the student and the teacher to increase comparing to other two diseases as shown in Fig.~\ref{roc3}.


\section{Related Work}

\label{Related}

Sevearal works  have been proposed in literature to study how machine learning can be used in medical diagnosis in the EHR domain. However, little works have addressed the problem of sharing the patients sensitive data.

In~\cite{choi2017generating}, Cho  \textit{et al.} have proposed a approach, called medical Generative Adversarial Network (medGAN), to generate
realistic synthetic patient records. Based on input real patient records, medGAN can generate high-dimensional discrete variables (e.g., binary and count features) via a combination of an autoencoder and generative adversarial
networks. However, the scheme shows privacy risks in both identity and attribute disclosure.

The use of mimic learning approach has been used in the \textit{information retrieval} (IR) domain, which a user can query an element of collection of huge number of documents. In IR domain, getting an access to large-scale datasets is crucial for designing effective IR systems. However, due to privacy issues, having access to such large-scale datasets is a real challenge. In~\cite{DBLP2}, a mimic learning scheme  have been proposed to train a  shareable model using two different techniques namely weak- and full-supervisions~\cite{dehghani2017neural}. Then, using the sharable model, it is easy to create large-scale datasets. Unfortunately, current research work has not yet studied the use of mimic learning in electronic heath records which comprises a domain specific environment including the patients medical records and it is not allowed by law to share it with others. \footnote{Many works have studied data security and privacy~\cite{baza2019b,baza2018blockchain,parkccnc,omar1,omar2,parksmarnet,yilmaz2019expansion,shafee2019mimic,baza2019detecting,baza2019blockchain,pazos2019privacy,firmware2,Lightride}} \footnote{Creating communication protocols for secure content delivery for networks of drones \cite{Delay-Analysis,8647532,8885505,8886101,8412262,amer2018optimizing,amer2019performance,amer2020caching,8756296,amer2019mobility,chaccour2019Reliability,7925123,7605066}.}

\section{Conclusions}
\label{Conclusion}

In this paper, we tackled the problem of sharing patients medical records using the mimic learning approach.  A knowledge transfer methodology has been proposed for enabling hospitals to share models model without the need to share the sensitive patients records. We have evaluated the mimic learning approach using extensive experiments with the help of three different datasets of patients diseases. The evaluation results indicate the teacher and student models have identical performance. Moreover, as the teacher model accuracy increases, the student model also follow the teacher. These results indicate that the mimic learning is successful in transferring the knowledge of the teacher model to a shareable model without the need to share the patients sensitive data. We believe that this work can incentive large hospitals to share models so others can make use of it to further improving people health.

\bibliographystyle{IEEEtran}
\bibliography{CC} 

\begin{thebibliography}{10}
\providecommand{\url}[1]{#1}
\csname url@samestyle\endcsname
\providecommand{\newblock}{\relax}
\providecommand{\bibinfo}[2]{#2}
\providecommand{\BIBentrySTDinterwordspacing}{\spaceskip=0pt\relax}
\providecommand{\BIBentryALTinterwordstretchfactor}{4}
\providecommand{\BIBentryALTinterwordspacing}{\spaceskip=\fontdimen2\font plus
\BIBentryALTinterwordstretchfactor\fontdimen3\font minus
  \fontdimen4\font\relax}
\providecommand{\BIBforeignlanguage}[2]{{%
\expandafter\ifx\csname l@#1\endcsname\relax
\typeout{** WARNING: IEEEtran.bst: No hyphenation pattern has been}%
\typeout{** loaded for the language `#1'. Using the pattern for}%
\typeout{** the default language instead.}%
\else
\language=\csname l@#1\endcsname
\fi
#2}}
\providecommand{\BIBdecl}{\relax}
\BIBdecl

\bibitem{birkhead2015uses}
G.~S. Birkhead, M.~Klompas, and N.~R. Shah, ``Uses of electronic health records
  for public health surveillance to advance public health,'' \emph{Annual
  review of public health}, vol.~36, pp. 345--359, 2015.

\bibitem{shickel2017deep}
B.~Shickel, P.~J. Tighe, A.~Bihorac, and P.~Rashidi, ``Deep ehr: a survey of
  recent advances in deep learning techniques for electronic health record
  (ehr) analysis,'' \emph{IEEE journal of biomedical and health informatics},
  vol.~22, no.~5, pp. 1589--1604, 2017.

\bibitem{arndt2017tethered}
B.~G. Arndt, J.~W. Beasley, M.~D. Watkinson, J.~L. Temte, W.-J. Tuan, C.~A.
  Sinsky, and V.~J. Gilchrist, ``Tethered to the ehr: primary care physician
  workload assessment using ehr event log data and time-motion observations,''
  \emph{The Annals of Family Medicine}, vol.~15, no.~5, pp. 419--426, 2017.

\bibitem{rajkomar2018scalable}
A.~Rajkomar, E.~Oren, K.~Chen, A.~M. Dai, N.~Hajaj, M.~Hardt, P.~J. Liu,
  X.~Liu, J.~Marcus, M.~Sun \emph{et~al.}, ``Scalable and accurate deep
  learning with electronic health records,'' \emph{NPJ Digital Medicine},
  vol.~1, no.~1, p.~18, 2018.

\bibitem{ref:google}
\BIBentryALTinterwordspacing
Google machine learning project. [Online]. Available:
  \url{https://www.mercurynews.com/2017/03/03/google-computers-trained-to-detect-cancer/}
\BIBentrySTDinterwordspacing

\bibitem{ref:google2}
\BIBentryALTinterwordspacing
Google's deepmind. [Online]. Available:
  \url{https://deepmind.com/applied/deepmind-health/}
\BIBentrySTDinterwordspacing

\bibitem{obermeyer2016predicting}
Z.~Obermeyer and E.~J. Emanuel, ``Predicting the future big data, machine
  learning, and clinical medicine,'' \emph{The New England journal of
  medicine}, vol. 375, no.~13, p. 1216, 2016.

\bibitem{2}
\BIBentryALTinterwordspacing
Hipaa-general information. [Online]. Available:
  \url{http://www.cms.gov/HIPAAGenInfo/.}
\BIBentrySTDinterwordspacing

\bibitem{samarati2001protecting}
P.~Samarati, ``Protecting respondents identities in microdata release,''
  \emph{IEEE transactions on Knowledge and Data Engineering}, vol.~13, no.~6,
  pp. 1010--1027, 2001.

\bibitem{Cortes:1995:SN:218919.218929}
C.~Cortes and V.~Vapnik, ``Support-vector networks,'' \emph{Machine Learning},
  vol.~20, no.~3, pp. 273--297, Sep. 1995.

\bibitem{shruthi2019review}
U.~Shruthi, V.~Nagaveni, and B.~Raghavendra, ``A review on machine learning
  classification techniques for plant disease detection,'' in \emph{2019 5th
  International Conference on Advanced Computing \& Communication Systems
  (ICACCS)}.\hskip 1em plus 0.5em minus 0.4em\relax IEEE, 2019, pp. 281--284.

\bibitem{Breiman2001}
L.~Breiman, ``Random forests,'' \emph{Machine Learning}, vol.~45, no.~1, pp.
  5--32, Oct. 2001.

\bibitem{fredrikson2014privacy}
M.~Fredrikson, E.~Lantz, S.~Jha, S.~Lin, D.~Page, and T.~Ristenpart, ``Privacy
  in pharmacogenetics: An end-to-end case study of personalized warfarin
  dosing,'' in \emph{23rd $\{$USENIX$\}$ Security Symposium ($\{$USENIX$\}$
  Security 14)}, 2014, pp. 17--32.

\bibitem{UCI}
\BIBentryALTinterwordspacing
UCI, ``Breast cancer data set.'' [Online]. Available:
  \url{https://archive.ics.uci.edu/ml/datasets/Breast+Cancer+Wisconsin+(Diagnostic)}
\BIBentrySTDinterwordspacing

\bibitem{Kaggle}
\BIBentryALTinterwordspacing
Kaggle, ``Kaggle's cardiovascular disease data set.'' [Online]. Available:
  \url{https://www.kaggle.com/sulianova/cardiovascular-disease-dataset, last
  visited = 7/22/2019.}
\BIBentrySTDinterwordspacing

\bibitem{UCI2}
\BIBentryALTinterwordspacing
UCI, ``Heart disease data set.'' [Online]. Available:
  \url{https://archive.ics.uci.edu/ml/datasets/Heart+Disease, last visited =
  7/22/2019.}
\BIBentrySTDinterwordspacing

\bibitem{Han:2011:DMC:1972541}
J.~Han, M.~Kamber, and J.~Pei, \emph{Data Mining: Concepts and Techniques},
  3rd~ed.\hskip 1em plus 0.5em minus 0.4em\relax San Francisco, CA, USA: Morgan
  Kaufmann Publishers Inc., 2011.

\bibitem{choi2017generating}
E.~Choi, S.~Biswal, B.~Malin, J.~Duke, W.~F. Stewart, and J.~Sun, ``Generating
  multi-label discrete patient records using generative adversarial networks,''
  \emph{arXiv preprint arXiv:1703.06490}, 2017.

\bibitem{DBLP2}
M.~Dehghani, H.~Azarbonyad, J.~Kamps, and M.~de~Rijke, ``Share your model
  instead of your data: Privacy preserving mimic learning for ranking,'' in
  \emph{SIGIR 2017 Workshop on Neural Information Retrieval (Neu-IR'17)}, Aug.
  2017.

\bibitem{dehghani2017neural}
M.~Dehghani, H.~Zamani, A.~Severyn, J.~Kamps, and W.~B. Croft, ``Neural ranking
  models with weak supervision,'' in \emph{Proceedings of the 40th
  International ACM SIGIR Conference on Research and Development in Information
  Retrieval}.\hskip 1em plus 0.5em minus 0.4em\relax ACM, 2017, pp. 65--74.

\bibitem{baza2019b}
\textcolor{white}{M.~Baza, N.~Lasla, M.~Mahmoud, G.~Srivastava, and M.~Abdallah, ``B-ride: Ride
  sharing with privacy-preservation, trust and fair payment atop public
  blockchain,'' \emph{arXiv preprint arXiv:1906.09968}, 2019.}

\bibitem{baza2018blockchain}
\textcolor{white}{M.~Baza, M.~Nabil, N.~Lasla, K.~Fidan, M.~Mahmoud, and M.~Abdallah,
  ``Blockchain-based firmware update scheme tailored for autonomous vehicles,''
  \emph{arXiv preprint arXiv:1811.05905}, 2018.}

\bibitem{parkccnc}
\textcolor{white}{W.~Al~Amiri, M.~Baza, M.~Mahmoud, W.~Alasmary, and K.~Akkaya, ``Towards secure
  smart parking system using blockchain technology,'' \emph{Proc. of 17th IEEE
  Annual Consumer Communications $\&$ Networking Conference (CCNC), Las vegas,
  USA}, 2020.}

\bibitem{omar1}
\textcolor{white}{M.~Baza, M.~M. Fouda, A.~S.~T. Eldien, and H.~A. Mansour, ``An efficient
  distributed approach for key management in microgrids,'' \emph{Proc. of the
  Computer Engineering Conference (ICENCO), Egypt}, pp. 19--24, 2015.}

\bibitem{omar2}
\textcolor{white}{M.~Baza, M.~Fouda, M.~Nabil, A.~S. Tag, H.~Mansour, and M.~Mahmoud,
  ``Blockchain-based distributed key management approach tailored for smart
  grid,'' in \emph{Combating Security Challenges in the Age of Big Data}.\hskip
  1em plus 0.5em minus 0.4em\relax Springer, 2019.}

\bibitem{parksmarnet}
\textcolor{white}{W.~Al~Amiri, M.~Baza, M.~Mahmoud, K.~Banawan, W.~Alasmary, and K.~Akkaya,
  ``Privacy-preserving smart parking system using blockchain and private
  information retrieval,'' \emph{Proc. of the IEEE International Conference on
  Smart Applications, Communications and Networking (SmartNets 2019)}, 2020.}

\bibitem{yilmaz2019expansion}
\textcolor{white}{I.~Yilmaz and R.~Masum, ``Expansion of cyber attack data from unbalanced
  datasets using generative techniques,'' \emph{arXiv preprint
  arXiv:1912.04549}, 2019.}

\bibitem{shafee2019mimic}
\textcolor{white}{A.~Shafee, M.~Baza, D.~A. Talbert, M.~M. Fouda, M.~Nabil, and M.~Mahmoud,
  ``Mimic learning to generate a shareable network intrusion detection model,''
  \emph{Proc. of the IEEE Consumer Communications \& Networking Conference,Las
  Vegas, USA}, 2020.}

\bibitem{baza2019detecting}
\textcolor{white}{M.~Baza, M.~Nabil, N.~Bewermeier, K.~Fidan, M.~Mahmoud, and M.~Abdallah,
  ``Detecting sybil attacks using proofs of work and location in vanets,''
  \emph{arXiv preprint arXiv:1904.05845}, 2019.}

\bibitem{baza2019blockchain}
\textcolor{white}{M.~Baza, M.~Nabil, M.~Ismail, M.~Mahmoud, E.~Serpedin, and M.~Rahman,
  ``Blockchain-based charging coordination mechanism for smart grid energy
  storage units,'' \emph{Proc. Of IEEE International Conference on Blockchain,
  Atlanta, USA}, July, 2019.}

\bibitem{pazos2019privacy}
\textcolor{white}{M.~Baza, M.~Pazos-Revilla, M.~Nabil, A.~Sherif, M.~Mahmoud, and W.~Alasmary,
  ``Privacy-preserving and collusion-resistant charging coordination schemes
  for smart grid,'' \emph{arXiv preprint arXiv:1905.04666}, 2019.}

\bibitem{firmware2}
\textcolor{white}{M.~Baza, J.~Baxter, N.~Lasla, M.~Mahmoud, M.~Abdallah, and M.~Younis,
  ``Incentivized and secure blockchain-based firmware update and dissemination
  for autonomous vehicles,'' in \emph{Connected and Autonomous Vehicles in
  Smart Cities}.\hskip 1em plus 0.5em minus 0.4em\relax CRC press, 2020.}

\bibitem{Lightride}
\textcolor{white}{M.~Baza, M.~Mahmoud, G.~Srivastava, W.~Alasmary, and M.~Younis, ``A light
  blockchain-powered privacy-preserving organization scheme for ride sharing
  services,'' \emph{Proc. of the IEEE 91th Vehicular Technology Conference
  (VTC-Spring), Antwerp, Belgium}, May 2020.}

\bibitem{Delay-Analysis}
\textcolor{white}{R.~Amer, M.~M. Butt, M.~Bennis, and N.~Marchetti, ``Delay analysis for wireless
  {D2D} caching with inter-cluster cooperation,'' in \emph{IEEE Global
  Communications Conference ({GLOBECOM})}, Singapore, Dec. 2017.}

\bibitem{8647532}
\textcolor{white}{R.~{Amer}, M.~M. {Butt}, H.~{ElSawy}, M.~{Bennis}, J.~{Kibilda}, and
  N.~{Marchetti}, ``On minimizing energy consumption for {D2D} clustered
  caching networks,'' in \emph{IEEE Global Communications Conference
  ({GLOBECOM})}, December 2018, pp. 1--6.}

\bibitem{8885505}
\textcolor{white}{R.~{Amer}, W.~{Saad}, H.~{ElSawy}, M.~M. {Butt}, and N.~{Marchetti}, ``Caching
  to the sky: Performance analysis of cache-assisted {CoMP} for
  cellular-connected {UAVs},'' in \emph{IEEE Wireless Communications and
  Networking Conference ({WCNC})}, April 2019, pp. 1--6.}

\bibitem{8886101}
\textcolor{white}{R.~{Amer}, H.~{ElSawy}, J.~{Kibi{\l}da}, M.~M. {Butt}, and N.~{Marchetti},
  ``Cooperative transmission and probabilistic caching for clustered {D2D}
  networks,'' in \emph{IEEE Wireless Communications and Networking Conference
  ({WCNC})}, April 2019, pp. 1--6.}

\bibitem{8412262}
\textcolor{white}{R.~Amer, M.~M. Butt, M.~Bennis, and N.~Marchetti, ``Inter-cluster cooperation
  for wireless {D2D} caching networks,'' \emph{IEEE Transactions on Wireless
  Communications}, vol.~17, no.~9, pp. 6108--6121, September 2018.}

\bibitem{amer2018optimizing}
\textcolor{white}{R.~Amer, H.~ElSawy, M.~M. Butt, E.~A. Jorswieck, M.~Bennis, and N.~Marchetti,
  ``Optimizing joint probabilistic caching and communication for clustered
  {D2D} networks,'' \emph{arXiv preprint arXiv:1810.05510}, 2018.}

\bibitem{amer2019performance}
\textcolor{white}{R.~Amer, H.~Elsawy, J.~Kibi{\l}da, M.~M. Butt, and N.~Marchetti, ``Performance
  analysis and optimization of cache-assisted {CoMP} for clustered {D2D}
  networks,'' \emph{submitted to IEEE Transactions on Mobile Computing}, 2019.}

\bibitem{amer2020caching}
\textcolor{white}{R.~Amer, W.~Saad, H.~ElSawy, M.~Butt, and N.~Marchetti, ``Caching to the sky:
  Performance analysis of cache-assisted {CoMP} for cellular-connected
  {UAVs},'' in \emph{Proc. of the IEEE Wireless Communications and Networking
  Conference ({WCNC})}, Marrakech, Morocco, April. 2019.}

\bibitem{8756296}
\textcolor{white}{R.~{Amer}, W.~{Saad}, and N.~{Marchetti}, ``Towards a connected sky:
  Performance of beamforming with down-tilted antennas for ground and {UAV}
  user co-existence,'' \emph{IEEE Communications Letters}, pp. 1--1, 2019.}

\bibitem{amer2019mobility}
\textcolor{white}{R.~Amer, W.~Saad, and N.~Marchetti, ``Mobility in the sky: Performance and
  mobility analysis for cellular-connected {UAVs},'' \emph{arXiv preprint
  arXiv:1908.07774}, 2019.}

\bibitem{chaccour2019Reliability}
\textcolor{white}{C.~Chaccour, R.~Amer, B.~Zhou, and W.~Saad, ``On the reliability of wireless
  virtual reality at terahertz ({THz}) frequencies,'' in \emph{10th IFIP
  International Conference on New Technologies}, Spain, June. 2019.}

\bibitem{7925123}
\textcolor{white}{R.~{Amer}, A.~A. {El-Sherif}, H.~{Ebrahim}, and A.~{Mokhtar}, ``Stability
  analysis for multi-user cooperative cognitive radio network with energy
  harvesting,'' in \emph{IEEE International Conference on Computer and
  Communications (ICCC)}, Oct 2016, pp. 2369--2375.}

\bibitem{7605066}
\textcolor{white}{R.~{Amer}, A.~A. {El-sherif}, H.~{Ebrahim}, and A.~{Mokhtar}, ``Cooperation and
  underlay mode selection in cognitive radio network,'' in \emph{International
  Conference on Future Generation Communication Technologies (FGCT)}, Aug 2016,
  pp. 36--41.}

\end{thebibliography}
    
\end{document}